\documentclass[10pt,conference,a4paper]{IEEEtran}
\IEEEoverridecommandlockouts

\ifCLASSINFOpdf
\else
\fi
\usepackage{subfigure} 
\usepackage{graphicx} 
\usepackage{hyperref}

\def\eop {{\noindent\framebox[0.5em]{\rule[0.25ex]{0em}{0.75ex}}}}
\def\BE{\begin{equation}} \def\EE{\end{equation}}
\def\BEA{\begin{eqnarray}} \def\EEA{\end{eqnarray}}

\begin{document}
%
\title{A Tight Convex Upper Bound on the Likelihood of a Finite Mixture}

\author{\IEEEauthorblockN{Elad Mezuman}
\IEEEauthorblockA{IBM Research - Haifa\\Email: eladme@il.ibm.com}
\and
\IEEEauthorblockN{Yair Weiss}
\IEEEauthorblockA{The Hebrew University of Jerusalem\\Email: yweiss@cs.huji.ac.il}
}


%


\maketitle

\begin{abstract}
The likelihood function of a finite mixture model is a non-convex
function with multiple local maxima and commonly used iterative
algorithms such as EM will converge to different solutions depending
on initial conditions. In this paper we ask: is it possible to assess how far we are from
the {\em global} maximum of the likelihood?

Since the likelihood of a finite mixture model can grow unboundedly by
centering a Gaussian on a single datapoint and shrinking the
covariance, we constrain the problem by assuming that the parameters
of the individual models are members of a  large discrete set
(e.g. estimating a mixture of two Gaussians where the means and
variances of both Gaussians are members of a set of a million possible
means and variances). For this setting we show that a simple upper
bound on the likelihood can be computed using convex optimization and
we analyze conditions under which the bound is guaranteed to be
tight. This bound can then be used to assess the quality of solutions
found by EM (where the final result is projected on the discrete set)
or any other mixture estimation algorithm. 
For any dataset our method allows us to find a finite
mixture model together with a dataset-specific bound on how far the
likelihood of this mixture is from the global optimum of the
likelihood.\end{abstract}


%
\IEEEpeerreviewmaketitle

\section{Introduction}
\begin{figure*}
\centerline{
\begin{tabular}{cccc}
Data & EM 1 & EM 2 & Our Bound\\
\subfigure[]{\includegraphics[width=0.23\textwidth]{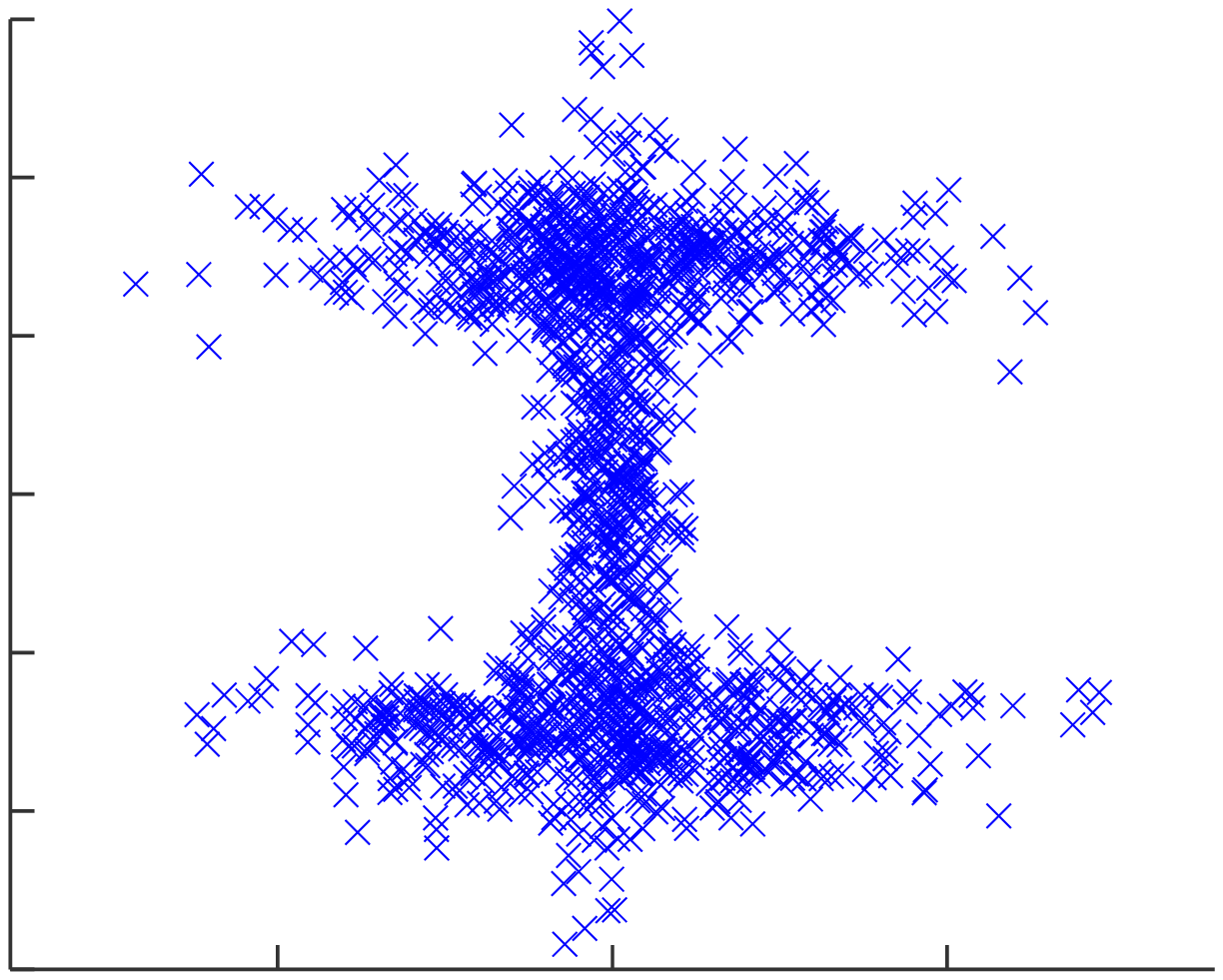}}
&
\subfigure[]{\includegraphics[width=0.23\textwidth]{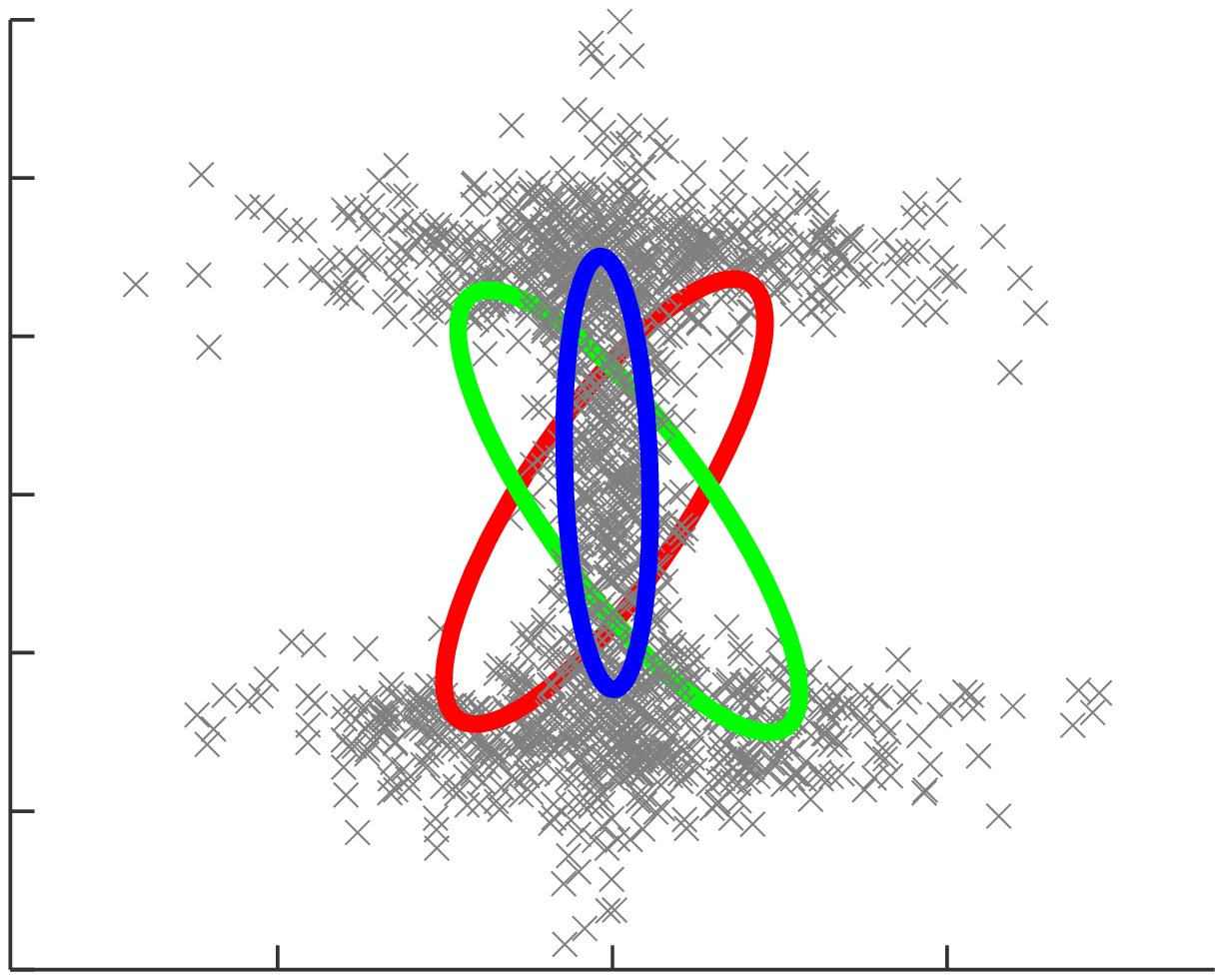}}
&
\subfigure[]{\includegraphics[width=0.23\textwidth]{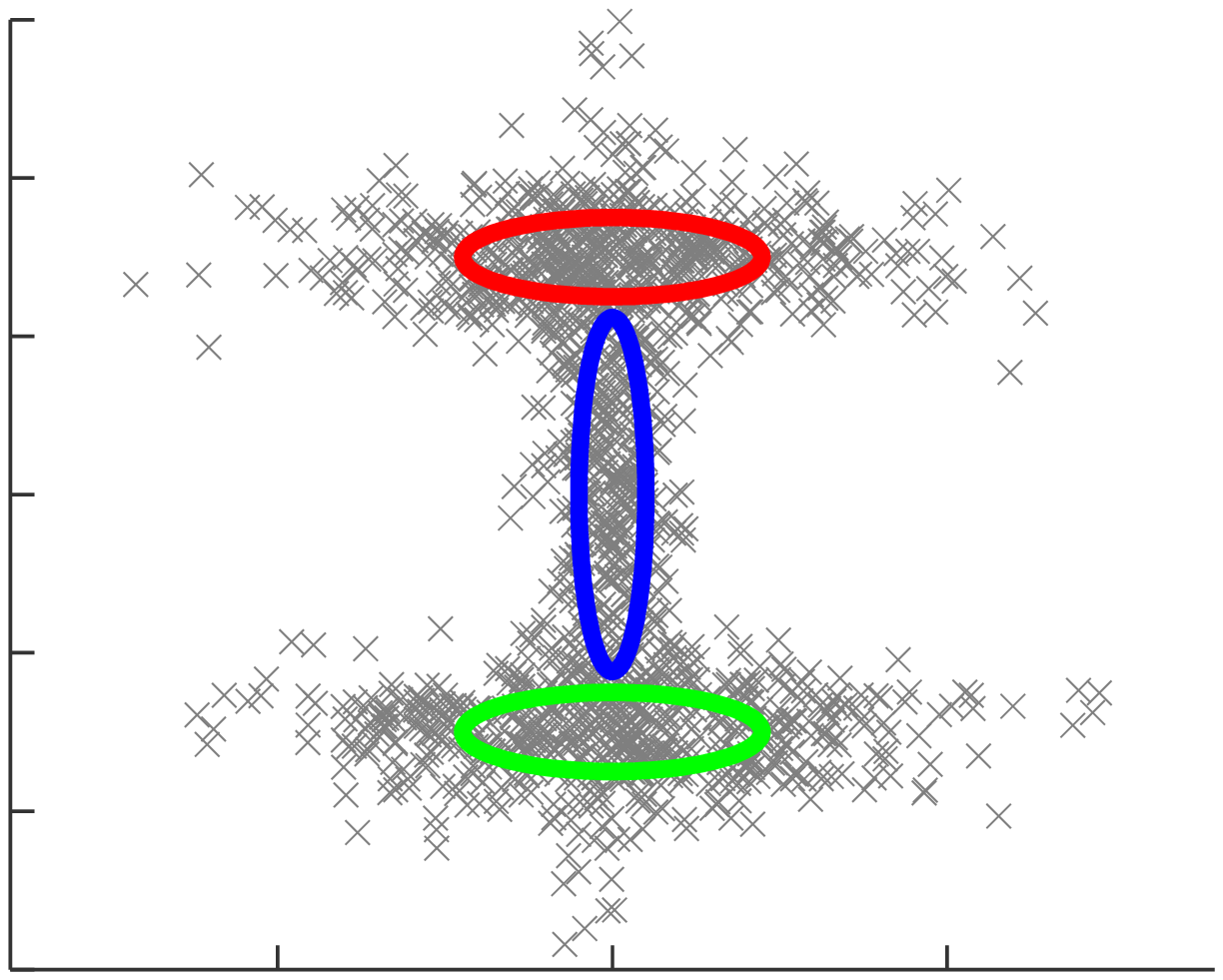}}
&
\subfigure[]{\includegraphics[width=0.23\textwidth]{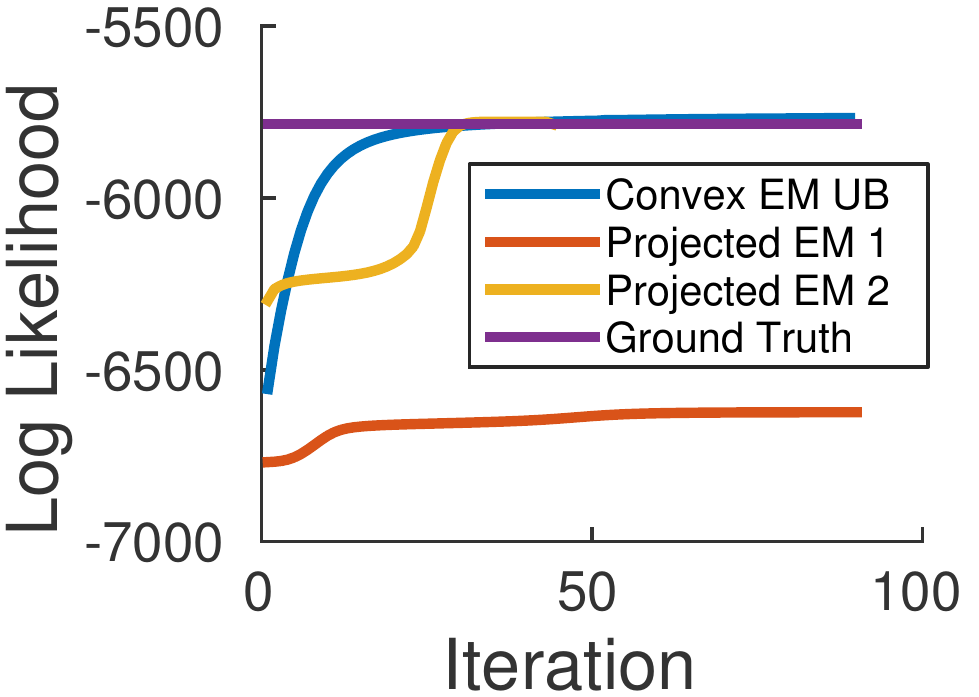}}
\end{tabular}
}
\caption{ {\bf a.} Simple synthetic dataset. {\bf b-c} Two possible solutions (depending on initial conditions) found by using EM for a mixture of three Gaussians. {\bf d.} We calculate an upper bound (Convex-EM) on the likelihood of any finite mixture whose components are in a large discrete set. Since the solution shown in {\bf d} achieves the upper bound, we can prove that it is the global optimum of the likelihood.}
\label{teaser-fig}
\end{figure*}
Finite mixture models are a primary workhorse of machine learning and pattern recognition and 
are frequently used in a wide range of
applications (e.g.~\cite{Machlachlan,grabcut,goldberger}). By far the
most common approach to estimating the parameters of a mixture model
is the Expectation-Maximization (EM) algorithm that iteratively
improves the likelihood of the data at every iteration~\cite{dempster1977maximum,Bishop}. The widespread
use of EM for solving such problems is despite its well-known dependence on initial conditions~\cite{Balakrishnan2014}. In fact the common practice is to rerun the algorithm
multiple times and choose the local maximum with highest
likelihood~\cite{Machlachlan}. Figure~\ref{teaser-fig} shows a simple
synthetic dataset similar to a problem set at Brown University
(\url{http://cs.brown.edu/courses/cs195-5}). Depending on the initial conditions, EM may converge to the
correct parameters (figure~\ref{teaser-fig}c) or very different
parameters (figure~\ref{teaser-fig}b).

In order to better understand the behavior of EM, a number of recent
papers have analyzed the case of EM for Gaussian mixture models when
the Gaussians are {\em well separated} and in high dimensions. For
such cases it can be shown that EM will converge rapidly to the
correct parameters~\cite{Dasgupta,Arora}. An alternative to EM are
moment-based methods which go back to the days of
Pearson~\cite{Pearson94} and seek parameters whose predicted moments
exactly match the empirical moments. Often the number of moments needed to be
matched grow rapidly with the size of the mixture and lead to
nonlinear equations. Recent spectral methods~\cite{HsuKakade} bypass many of
the problems of traditional moment methods but they require specific forms for the covariance. The data in
figure~\ref{teaser-fig} contains three elliptical Gaussians in two dimensions and
the densities of the Gaussians overlap significantly. Can we globally maximize the
likelihood for such data?

A trivial solution to maximizing the likelihood of a Gaussian mixture model is to set the mean of one of the Gaussians on a single datapoint and shrink the covariance to zero. This solution can increase the likelihood of the data arbitrarily~\cite{Dasgupta,Machlachlan}. Thus the problem of finding the global maximum only makes sense if we restrict the set of possible models. In this paper, we adopt the restriction that each of the $K$ models must  belong to a large, discrete set of
$M$ possible parameters. Can we find the maximum likelihood solution within this constrained set of parameters? This strategy of restricting the search to a large, discrete set of possible parameters has been successful in the exemplar approach to clustering. Two prominent examples are  k-medoids~\cite{kaufman1987clustering} and affinity propagation~\cite{ap} which show that by using discrete optimization one can often obtain better clustering solutions compared to  k-means.  Lashkari and Golland~\cite{lashkari2007convex} presented a convex formulation of exemplar based clustering using a variant of EM. Nowozin and Gokhan extended this approach by allowing the algorithm to suggest additional exemplars beyond the training data~\cite{nowozin2008decoupled}. 

Even with the restriction that the model parameters belong to a
discrete set the problem is seemingly combinatorial: we still need to
find the $K$ ``active'' parameters out of the $M$ possibilities and
the mixing weights on each of the models. In this paper, we show that
a simple upper bound on the likelihood can be computed using convex
optimization and that this bound is guaranteed to be tight in the
generative setting as the number of data points goes to infinity. This bound can then be used to assess the quality of a solution achieved by any mixture estimation algorithm, provided it satisfies the constraints.
 Thus for any dataset our 
method allows us to find a finite mixture model together with a
dataset-specific bound on how far the likelihood of this mixture is
from the global optimum of the likelihood.

\section{Problem Formulation}

In a finite mixture model we consider models of the form:
\BE
\Pr(x;\pi,\theta)= \sum_{k=1}^K \pi_k \Pr(x;\theta_k)
\EE

Given a dataset $\{x_i\}$ we seek to find a vector of mixing weights of size $K$ and $K$ parameter vectors  to maximize the likelihood of the data.  In the constrained 
maximum likelihood problem we are also given a set of possible parameter vectors $\Theta=\{\theta_m\}_{m=1}^M$ and the problem becomes:
\BE
\label{constrained-ml-equation}
MLE=  \max_{\pi,\{m_l\}_{l=1}^K \subset \{1 \cdots M\}} \prod_i \sum_{k=1}^K \pi_k \Pr(x_i;\theta_{m_k})
\EE

\begin{figure*}
\begin{centering}
\subfigure[]{\includegraphics[width=0.20\textwidth]{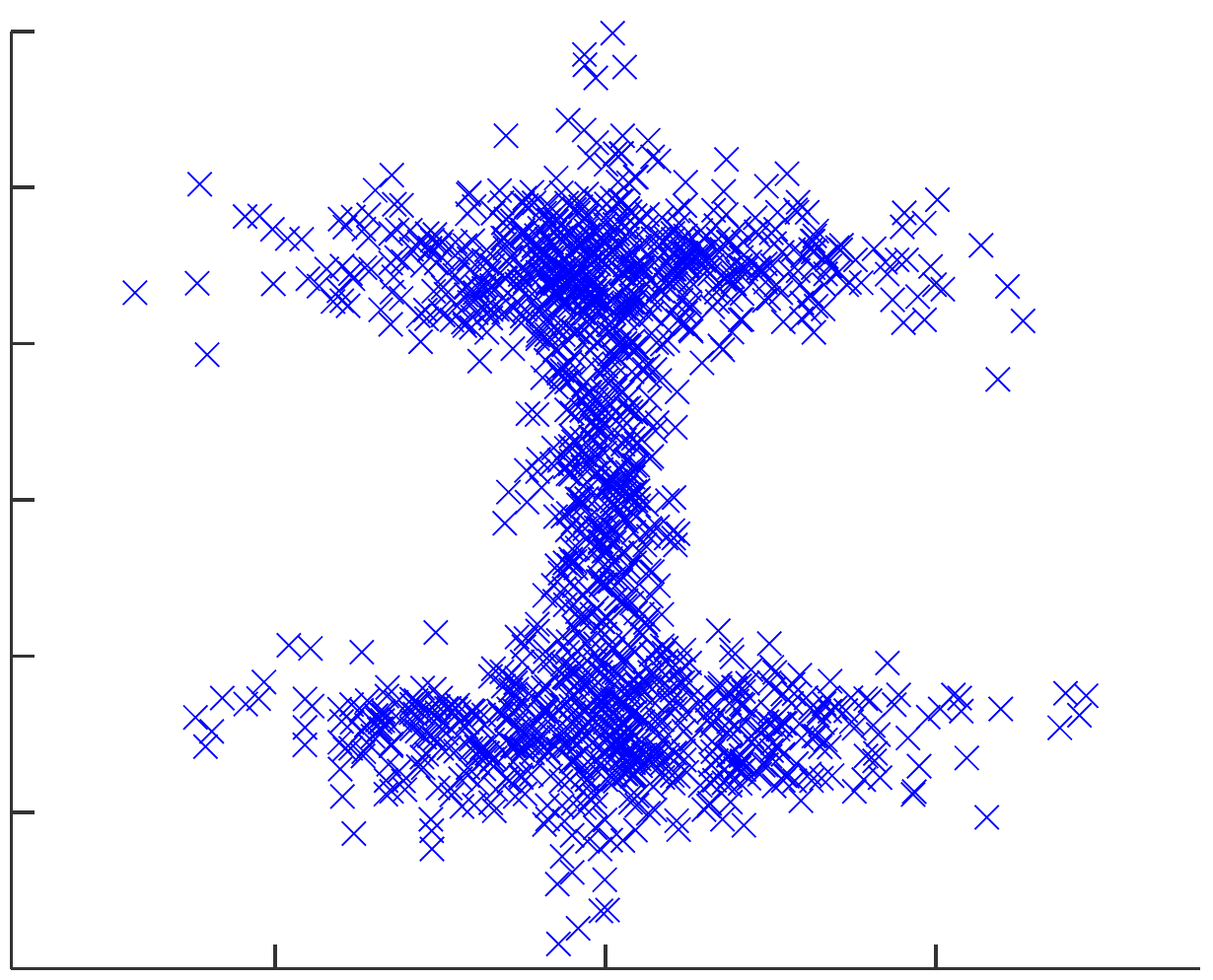}}
\hspace{40pt}
\subfigure[]{\includegraphics[width=0.21\textwidth]{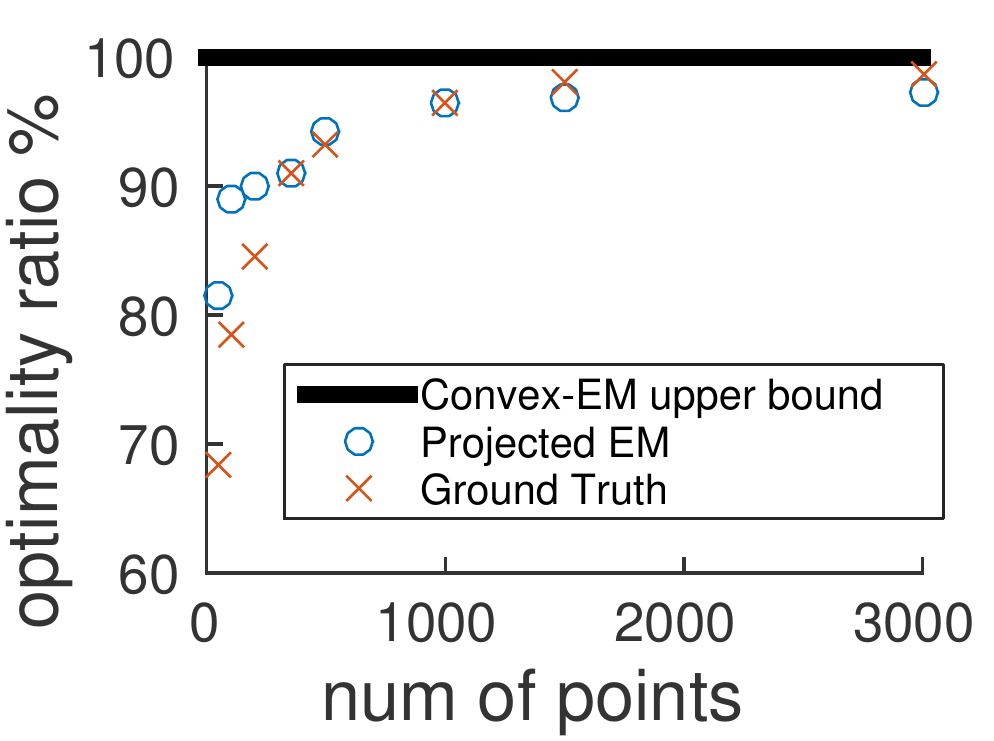}}
\hspace{40pt}
\subfigure[]{\includegraphics[width=0.20\textwidth]{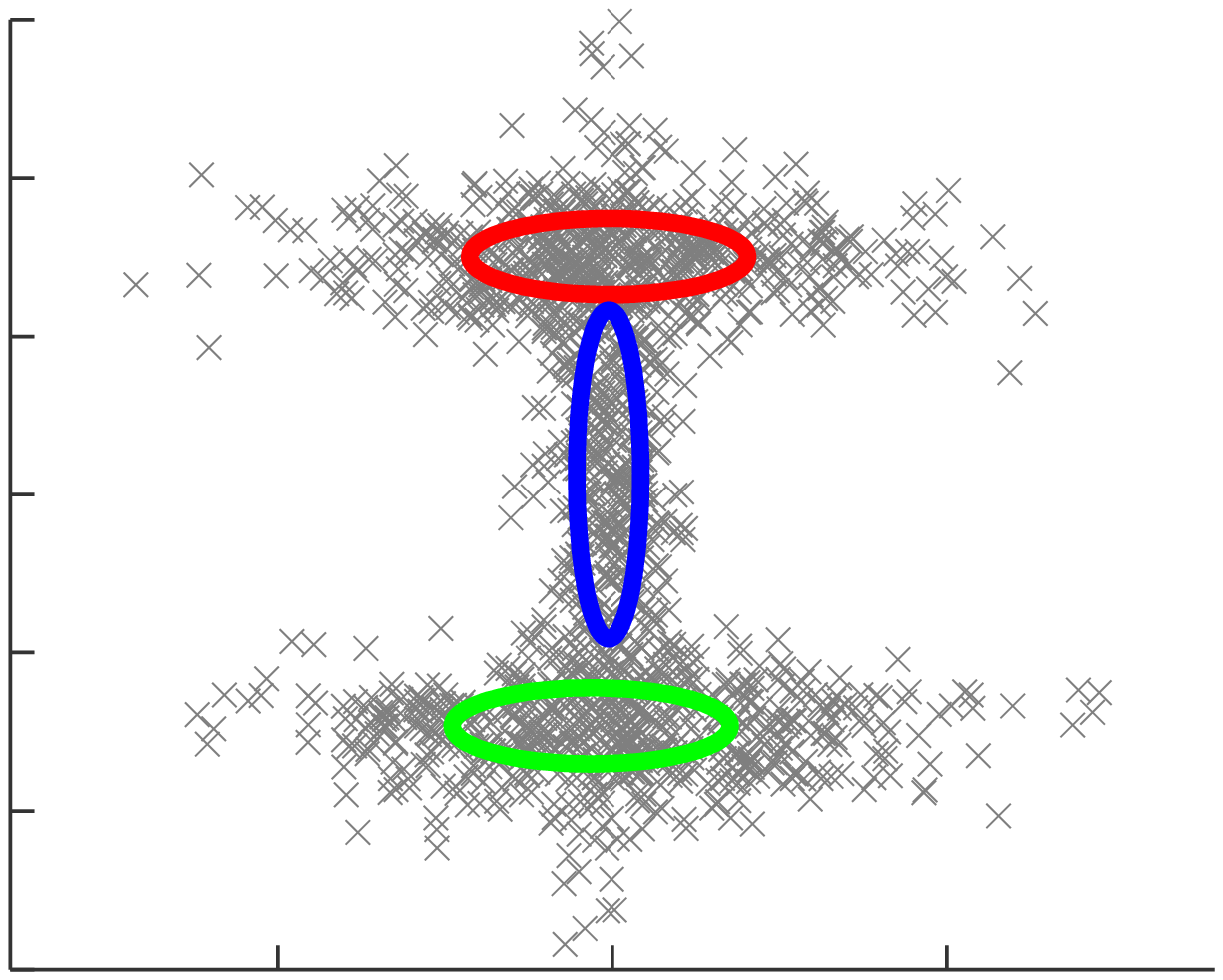}}
\\
\subfigure[]{\includegraphics[width=0.20\textwidth]{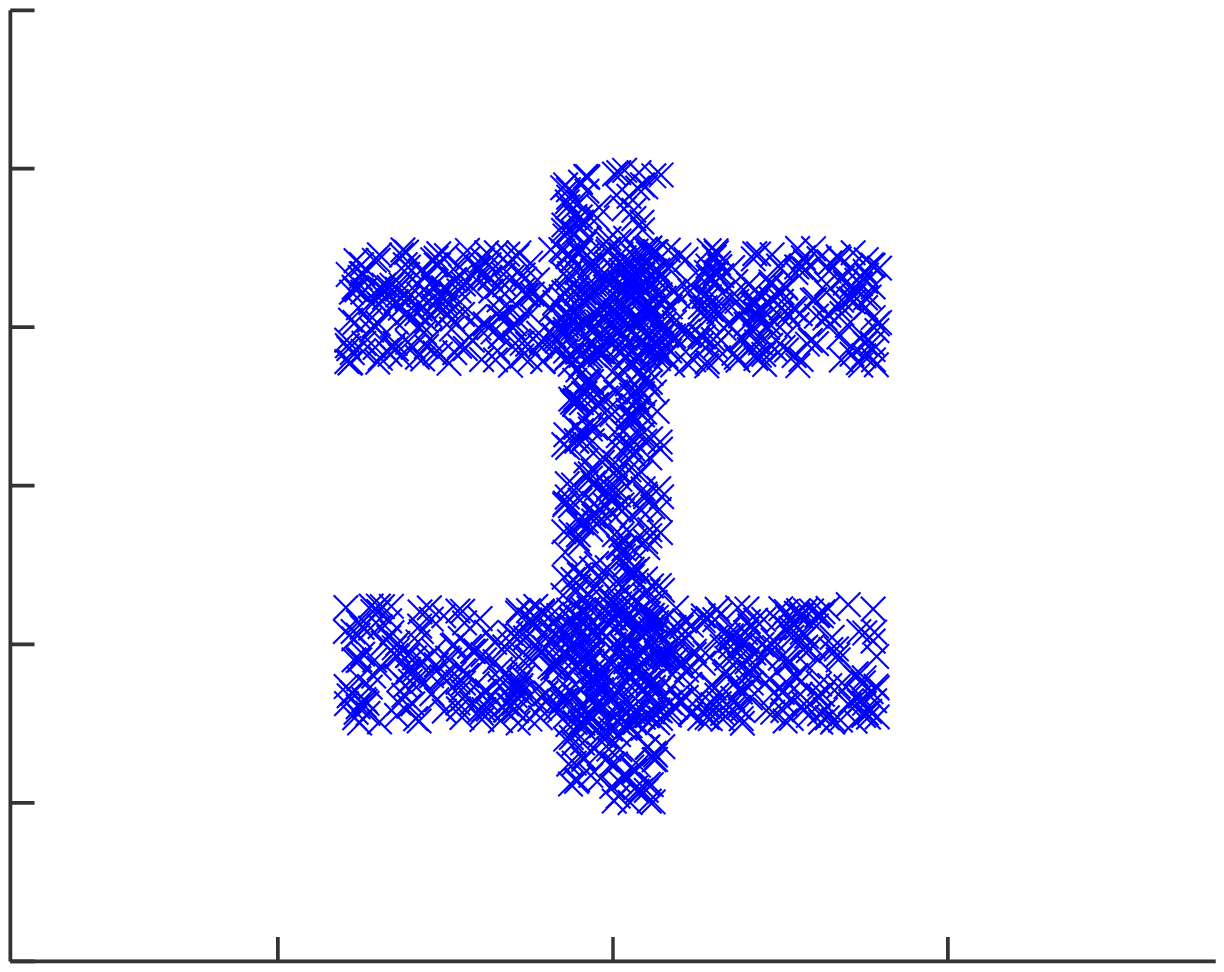}}
\hspace{40pt}
\subfigure[]{\includegraphics[width=0.22\textwidth]{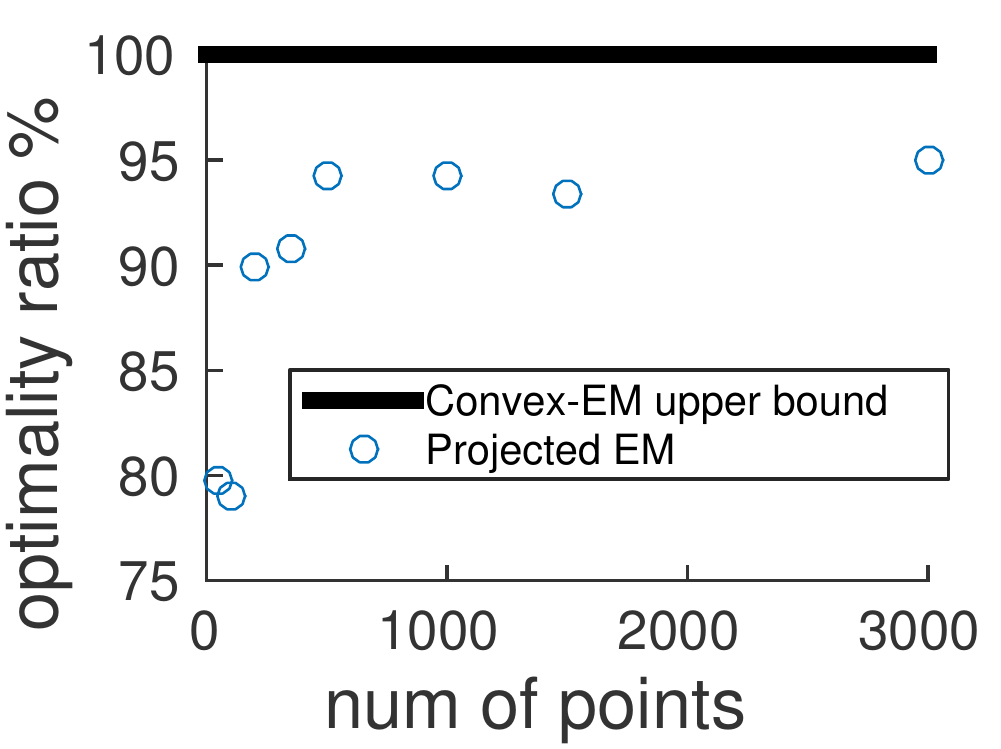}}
\hspace{40pt}
\subfigure[]{\includegraphics[width=0.20\textwidth]{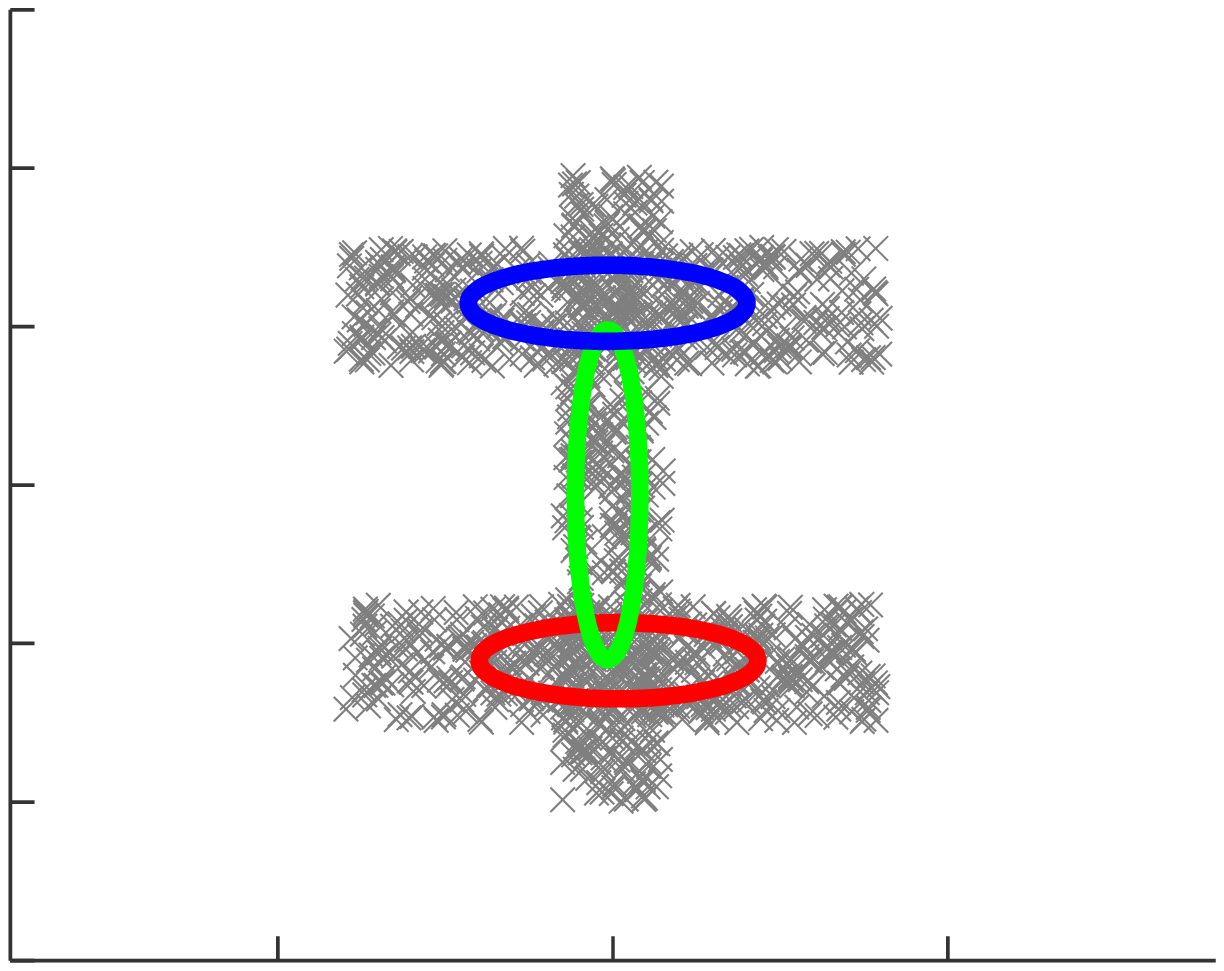}}
\par\end{centering}
\caption[]{{\bf a.} Synthetic data from a mixture of three Gaussians. {\bf b.} The ratio between the log likelihood of finite Gaussian mixture models (GMM) and the best GMM found using multiple restarts of EM (projected EM), and the upper bound found using convex optimization as a function of the number of points. This log likelihood  is ``calibrated'' by subtracting the average log likelihood of a random model to produce the ``optimality ratio''. The bound becomes tight as the number of points increases. {\bf c.} The GMM that achieves the best lower bound. {\bf d-f.} Same results with data that was not generated by a GMM and using exactly the same algorithm. Even though the data was not generated by a GMM the bound still tightens.}
\label{tightness-fig}
\end{figure*}

\section{An Upper Bound}
We can rewrite the constrained ML problem (equation~\ref{constrained-ml-equation}) as follows:
\BE
\label{constrained-ml-equation-sparse}
MLE=  \max_{\|\pi\|_0=K}  \prod_i \sum_{m=1}^M \pi_m \Pr(x_i;\theta_m)
\EE
Where we now represent the solution as a vector $\pi$ of length $M$ with only $K$ nonzero elements. This is obviously equivalent to the original MLE problem.  By relaxing the cardinality constraint we get an immediate upper bound:
\BE
\label{constrained-ml-equation-sparse-bound}
MLE \leq \max_{\pi}  \prod_i \sum_{m=1}^M \pi_m \Pr(x_i;\theta_m)
\EE

{\bf Observation 1:} The calculation of the upper bound (equation~\ref{constrained-ml-equation-sparse-bound}) is equivalent to solving a convex optimization problem.

{\bf Proof:} This result is well known(e.g.~\cite{HelmboldSSW95,lashkari2007convex}.)  and we sketch  the proof for completeness. We can rewrite the problem as:
\BE
UB=\max_{\pi} \sum_i \log \left( \sum_m \pi_m P_{x_i,m} \right) 
\EE
where $P_{x_i,m}$ is a fixed matrix. Direct calculation shows that this problem is concave.

{\bf Observation 2:} The calculation of the upper bound (equation~\ref{constrained-ml-equation-sparse-bound}) can be performed using a variant of the EM algorithm. For this variant, the solution does not depend on initialization.

{\bf Proof:}
This result is also well known(e.g.~\cite{HelmboldSSW95,lashkari2007convex}.) and we sketch the proof for completeness. It is known that at every iteration of EM the likelihood increases or stays the same, and that if EM converges it is a local maximum of the likelihood. Since for this problem the log likelihood is concave any local maximum is also globally optimal.

 The form of EM for this restricted setting is:

\BEA
\label{em-iterations1}
q_i(m)&\leftarrow& \frac{\pi_m \Pr(x_i;\theta_m)}{ \sum_l \pi_l \Pr(x_i;\theta_l)} \\
\label{em-iterations2}
\pi_m  & \leftarrow & \frac{1}{N}  \sum_i q_i(m) 
\EEA
 Helmbold et al.~\cite{HelmboldSSW95} show how to speed up the convergence of these iterations by using overrelaxation: $\pi^{new}=\pi^{old}+\eta (\pi^{new}-\pi^{old})$ with $\eta>1$. We refer to this algorithm (iterating equations~\ref{em-iterations1}-\ref{em-iterations2}) as the ``convex EM algorithm''.

To summarize, by removing the requirement that the mixture be a finite
mixture, or equivalently that the vector $\pi$ has only $K$
nonzero components, we obtain a concave problem that can be solved
using EM regardless of initial condition. But why should this bound be
tight? After all, the space of $K$ sparse probability vectors of length
$M$ is a vanishing fraction of the space of all probability vectors of
length $M$. It would therefore seem that by allowing a much larger search
space we will obtain higher log likelihood and thus the bound will
never be tight. In the next section, we show that this intuition is
wrong and in fact the bound will be tight under reasonable conditions.

\subsection{Tightness of Upper Bound}

We start with the {\em generative} setting. We assume we are given a dataset $\{x_i\}_{i=1}^N$ which are IID samples from a finite mixture density $p$. We also assume that we are given a class of possible components $\Theta$ and that all $K$ components of the generating distribution $p$ are members of the discrete set $\Theta$. We now show that under these assumptions the bound is guaranteed to be tight as the number of points $N$ goes to infinity and give a characterization of the looseness of the bound for finite $N$.

We denote by 
$$
p_\pi (x) =  \sum_{m=1}^M \pi_m \Pr(x;\theta_m)
$$

{\bf Definition:} The normalized log likelihood that a weight  vector $\pi$ gives to a dataset $\{x_i\}_{i=1}^N$  is denoted by $LL(\pi)= \frac{1}{N} \sum_i \log p_\pi(x_i)$

{\bf Definition:} For a dataset $\{x_i\}$ and a fixed set of models $\Theta$ we denote the upper bound computed by maximizing equation~\ref{constrained-ml-equation-sparse-bound} by:
\BE
UB= \max_{\pi} LL(\pi)
\EE

The main idea of the proof is to view the normalized log likelihood as an estimate of the negative cross-entropy between $p$ and the distribution $p_\pi$.   Clearly as the size of the dataset goes to infinity $LL(\pi)$ converges to the negative cross entropy $\int_x p(x) \log p_{\pi} (x) dx$. (Recall that for any function $f(x)$, $\frac{1}{N} \sum f(x_i)$ converges to $\int f(x) p(x) dx$).  To state the tightness of the upper bound we will need to characterize the worst-case difference between the log likelihood $LL(\pi)$ and the negative cross entropy between $p$ and $p_\pi$.

{\bf Definition:} The  cross entropy estimation error $EE(p,N)$ is defined as:
\BEA 
\nonumber EE(p,N) &=\\
\nonumber \max_{\pi} \left| \int_x p(x) \log p_\pi(x) dx - \frac{1}{N} \sum_i \log p_{\pi}(x_i) \right| &= \\
    \max_{\pi} \left| \int_x p(x) \log p_\pi(x) dx - LL(\pi) \right|
\EEA

Standard results (e.g.~\cite{kay1998fundamentals}) show that for any $p$ $EE(p,N)$ will approach zero as $N$ goes to infinity (since $LL$ is a sample average and the negative cross entropy is its expected value, the expected squared distance between them decreases as $1/N$).

We can now state our main result.

{\bf Theorem 1:} In the generative setting, let $\pi^*$ be a $K$ sparse vector that represents the generative   distribution $p=p_{\pi^*}=\sum_m \pi^*_m \Pr(x;\theta_m)$ then 
\BE
UB(N) \leq LL(\pi^*) + 2 EE(p,N)
\EE
In particular as $N$ goes to infinity the bound is tight and $LL(\pi^*)=UB(N)$.

{\bf Proof:} 
\BEA
UB &=&  \max_{\pi} LL(\pi) \\
\nonumber
&=&    \max_{\pi}   LL(\pi) -\int_x p(x) \log p_\pi(x) dx\\
&&  + \int_x p(x) \log p_\pi(x) dx  \\
\nonumber
& \leq&   \max_{\pi}   \left( LL(\pi) -\int_x p(x) \log p_\pi(x) dx \right) \\
 && + \max_{\pi} \int_x p(x) \log p_\pi(x) dx \\
&\leq & EE(p,N)+ \int_x p(x) \log p(x) dx
\label{rhs-eq}
\EEA
where we have used the fact that the cross entropy between any two distributions $p_1,p_2$ is minimal when $p_1=p_2$. Now, by the definition of the entropy estimation error $EE$ we also know that 
$|LL(\pi^*)-\int_x p(x) \log p(x) dx| \leq EE(p,N)$. Substituting into the right hand side of equation~\ref{rhs-eq} shows that $UB \leq LL(\pi^*) + 2 EE(p,N)$. \eop

Figure~\ref{tightness-fig} illustrates the result in theorem 1
for the data shown in Figure~\ref{teaser-fig} where the class of
models contains more than a million different models. Even though the upper bound is optimized over
all possible probability vectors without any explicit
sparsity penalty, the bound becomes tight as the amount of data grows
and already at $100$ points per cluster the true model achieves about 98\%
of the upper bound.

What happens when the data is not generated by a finite mixture whose
components are in the discrete set $\Theta$? It can be shown that the
tightness of the bound degrades gracefully: as long as there exists a
finite mixture which is close to the generating distribution (i.e. has KL
divergence at most $\epsilon$ from the generating distribution) then
as the number of points $N$ goes to infinity the bound will be at most
$\epsilon$ away from the log likelihood of the close finite mixture. 
We formalize this in the following theorem.

{\bf Theorem 2:} let $\pi^*$ be a $K$ sparse vector such that $p_{\pi^*}$  is closest to the generating distributing $p$ and  $KL(p||p_{\pi^*})=\epsilon$ then:
\BE
UB(N) \leq LL(\pi^*) + 2 EE(p,N) + \epsilon
\EE
In particular as $N$ goes to infinity the gap between the bound and what can be achieved by a finite mixture in the discrete set approaches $\epsilon$.

{\bf Proof:} This follows by substituting the fact that the KL between $p$ and $p_\pi$ is at most $\epsilon$ into equation~\ref{rhs-eq} and again using the definition of $EE(p,N)$.

Figure~\ref{tightness-fig}d-f illustrates this theorem. Here the data is generated by a mixture of three rectangles while the class of models $\Theta$ is the same class of a million Gaussians. Even though there is no way to represent the generating distribution as a mixture of three Gaussians, the upper bound that is computed using convex optimization becomes tighter as the number of points increases and already at $N=500$ there exist a mixture of three Gaussians that attains more than 98\% of the upper bound.

We note some properties of theorems 1 and 2.
\begin{itemize}
\item Theorem 1 shows that for any fixed set of models $\Theta$ the true generating distribution $p$
  optimizes the constrained likelihood problem as the number of points
  goes to infinity. This should be contrasted with unconstrained
  maximum likelihood estimation for Gaussian mixture models where by
  shrinking the variances arbitrarily we can achieve higher likelihood
  than the true parameters. 
\item The asymptotic tightness of the bound calculated using convex
  optimization does not depend on how ``well separated'' the mixture
  components are. 
\end{itemize}

\section{Obtaining a Lower Bound}

Theorems 1 and 2 show that we can recover an asymptotically tight
upper bound on the likelihood of the best sparse weight vector $\pi$
using convex optimization, i.e. by running the convex EM
algorithm. This bound can then be used to assess the quality of any algorithm that estimates a mixture model from data, provided the algorithm's output satisfies the constraint that all models come from the large, discrete set of models. 

The simplest algorithm for estimating a finite mixture that satisfies
the constraints is {\bf projected EM}.  We simply run the standard EM
algorithm and then ``project'' the result into the constraint set:
replace each model from the mixture found by EM with the closest model
in the discrete set. As shown by~\cite{Balakrishnan2014}, if EM is
initialized close to the correct solution it will converge to that
solution and so we use the standard method of using multiple random
restarts and choose the EM solution that gives highest likelihood after projection.

If the log likelihood found by projected EM (or any other algorithm) equals the upper bound then we have provably found the MLE. In practice, we need to define a numerical threshold that measures how far the lower bound is from the upper bound. We do this by defining the {\em optimality ratio}. this is simply the ratio between the log likelihood of a discrete solution and the upper bound. Since the ratio can be affected by adding a constant to the log likelihood, we ``calibrate'' the log likelihood by subtracting from it the average log likelihood of a random mixture model for this data. 
\[
opt\_ratio(\pi) = \frac{LL(\pi)-LL_{rand}}{UB-LL_{rand}}
\]
where $LL_{rand}$ is the average value of a random $\pi$ vector with cardinality $K$.
Thus an optimality ratio of zero means that the proposed solution is no better than a random $\pi$ vector and an optimality ratio of one means that the proposed solution is optimal.

\section{Experiments}
\begin{figure*}
\centerline{
\begin{tabular}{ccc}
Ground truth & Opt-Ratio=90\% & Opt-Ratio=96\%\\
\includegraphics[width=0.24\textwidth]{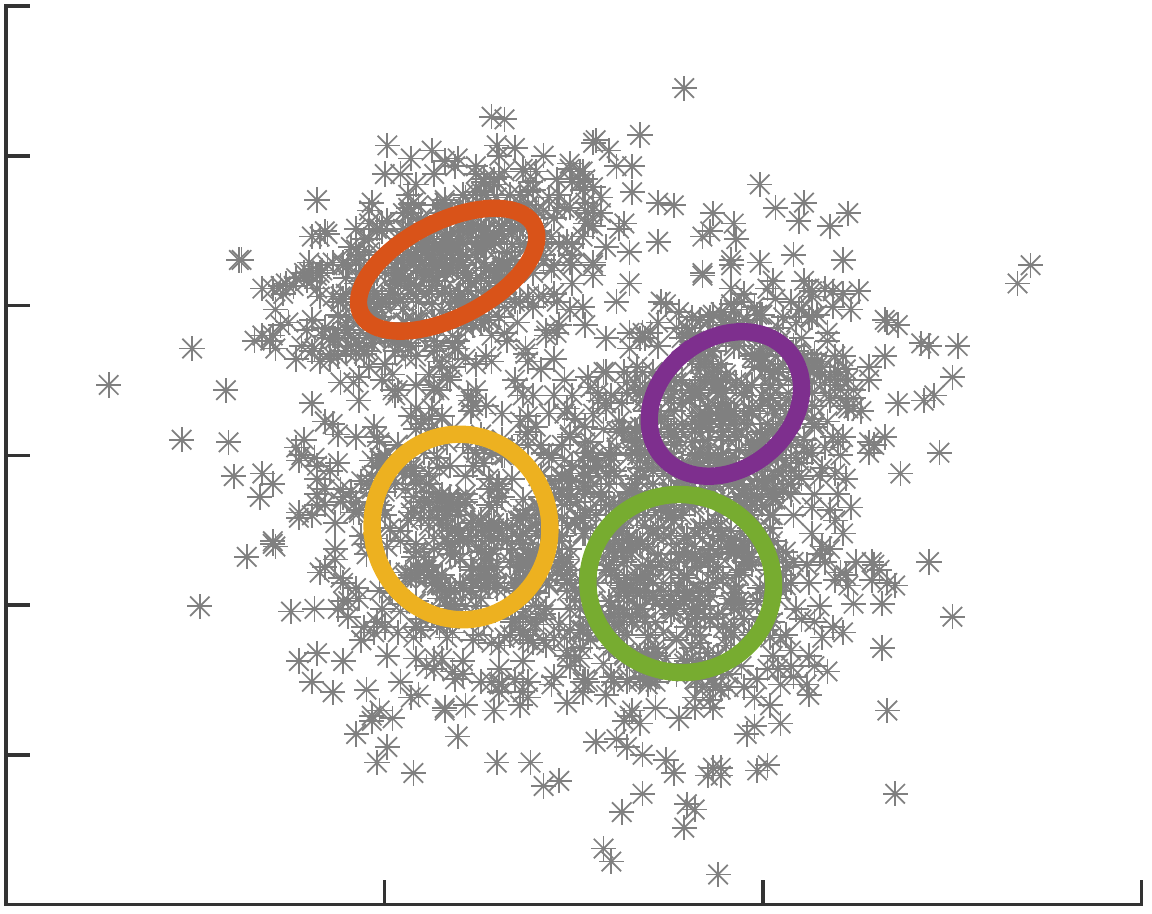} &
\includegraphics[width=0.24\textwidth]{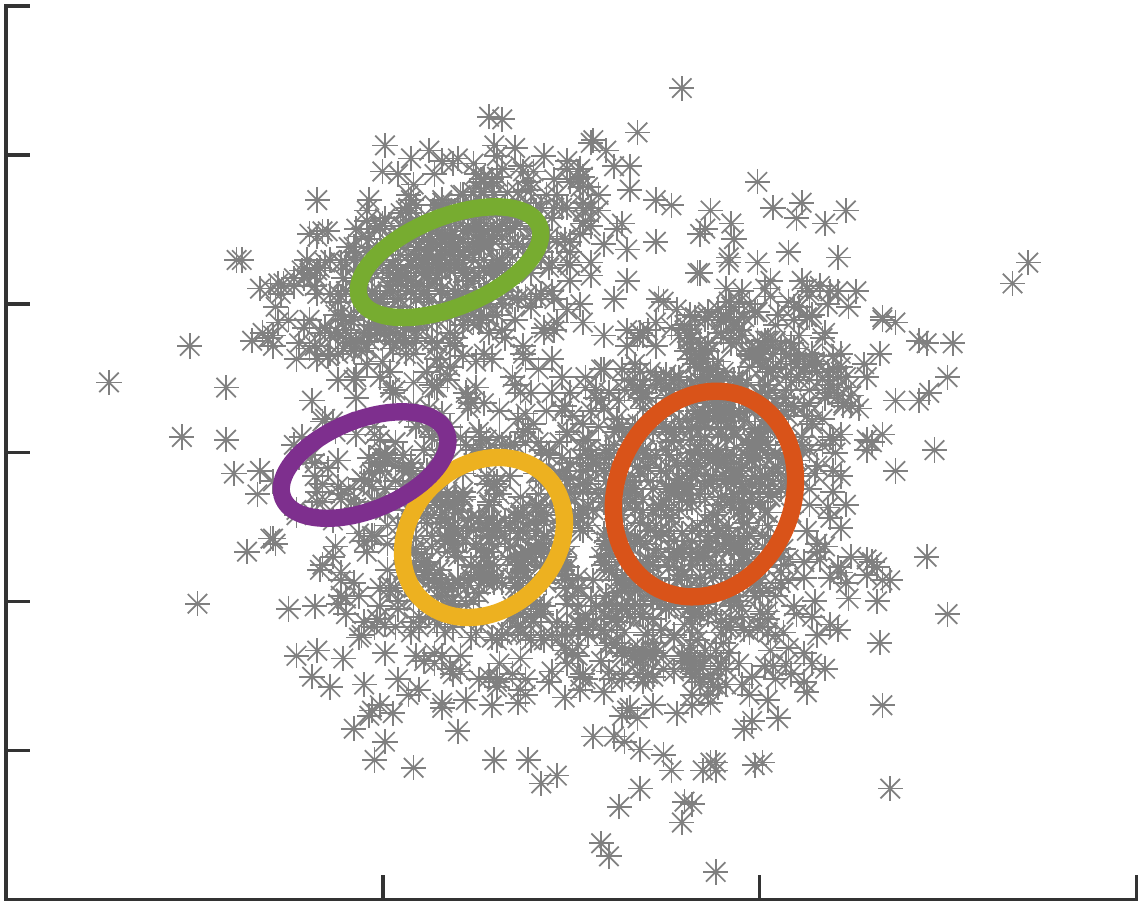} &
\includegraphics[width=0.24\textwidth]{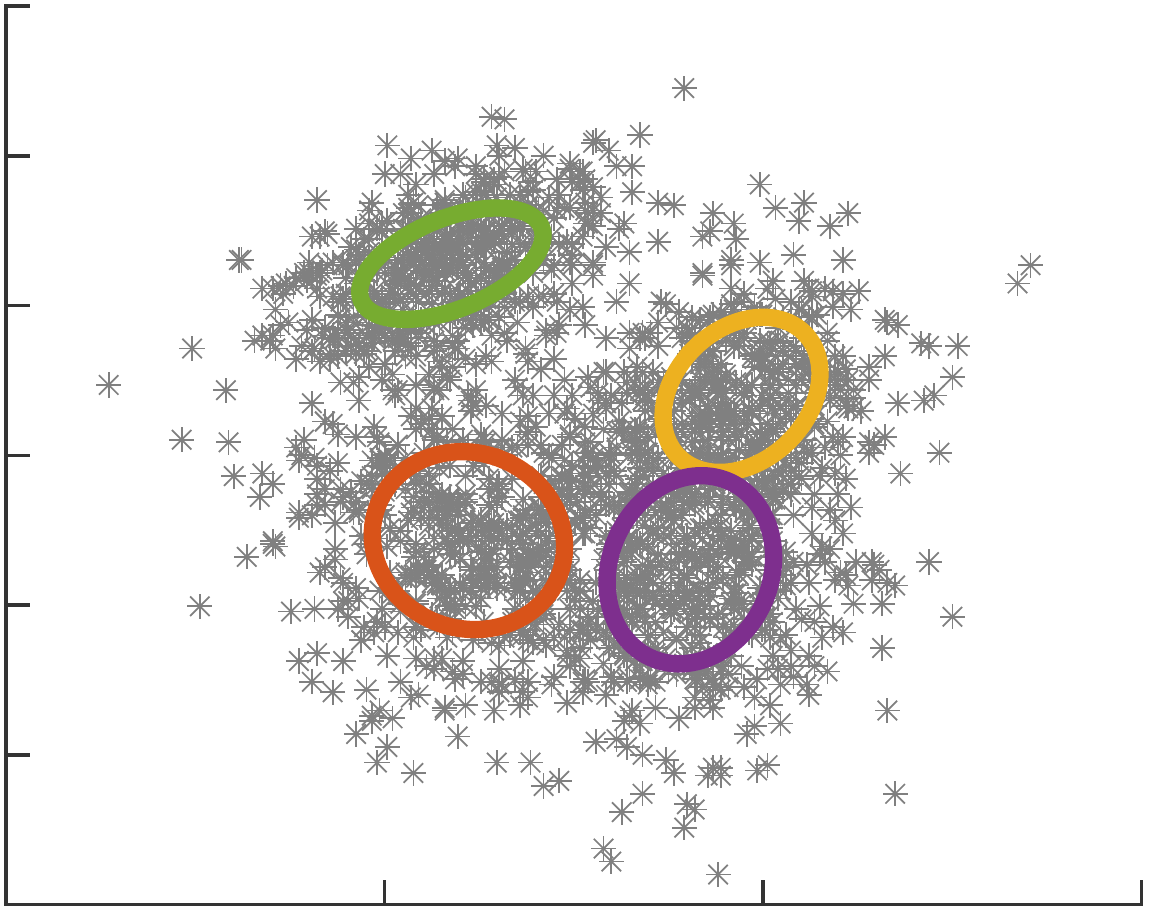}
\\ (a) & (b) & (c) \\
Discrete Set &  & \\
\includegraphics[width=0.24\textwidth]{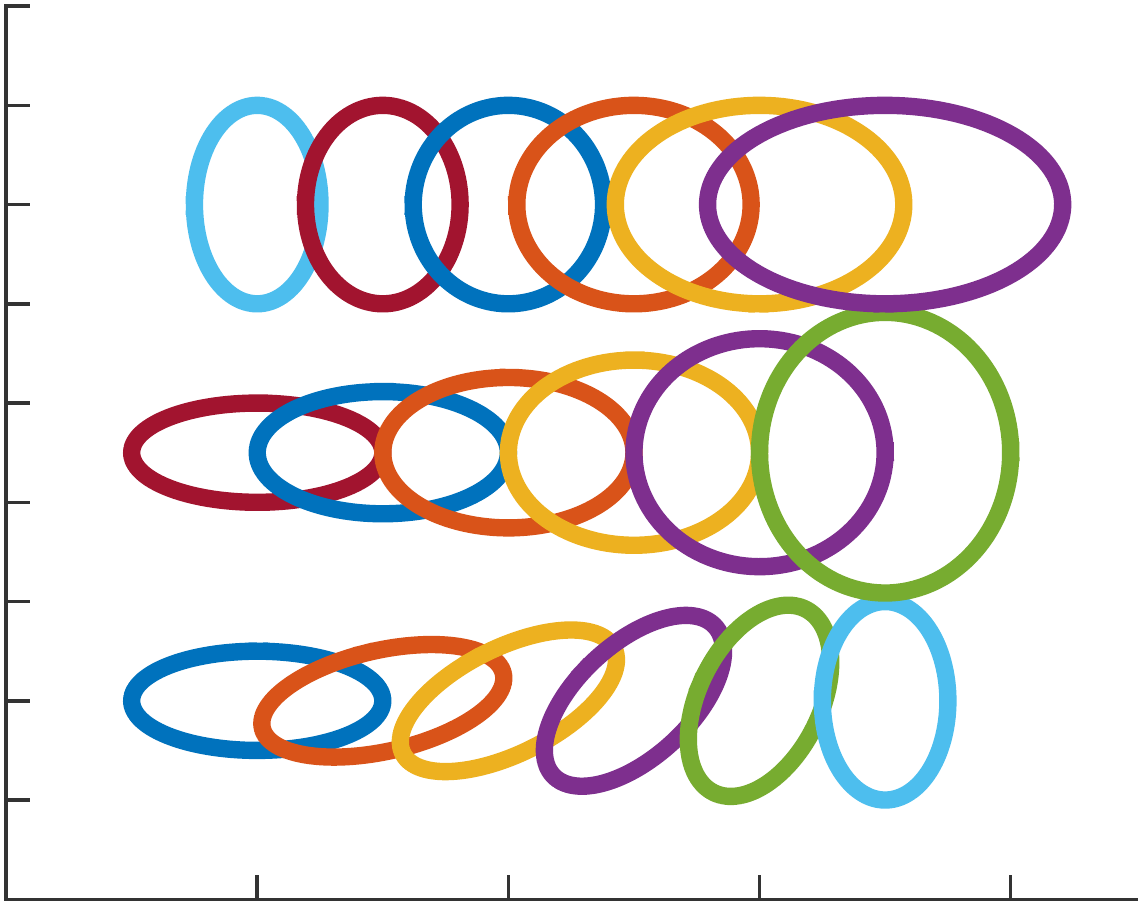} &
\includegraphics[width=0.24\textwidth]{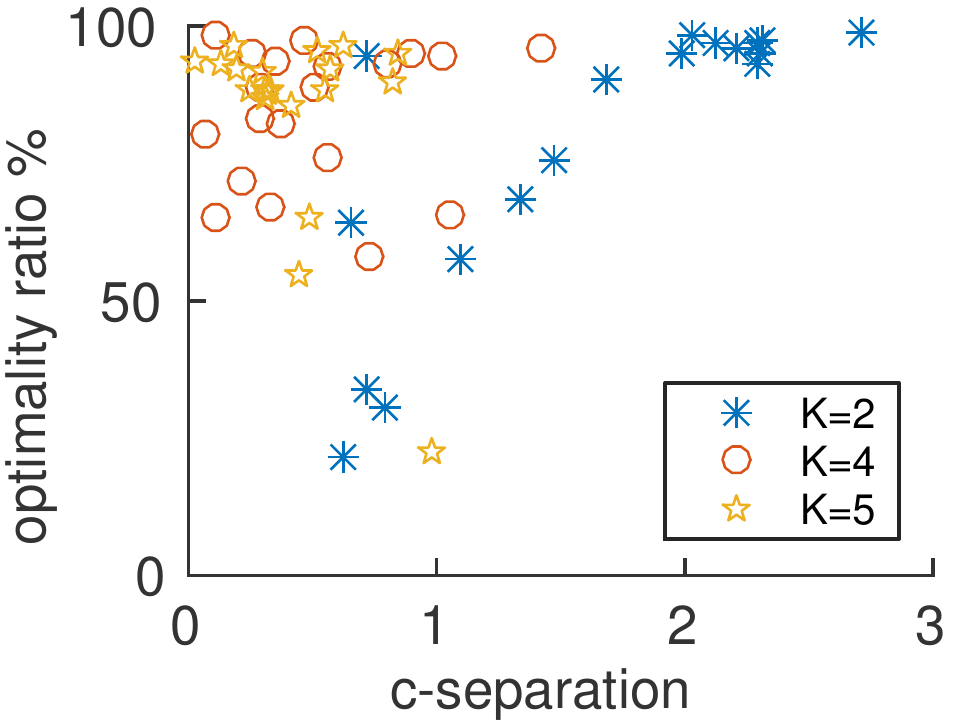} &
\includegraphics[width=0.24\textwidth]{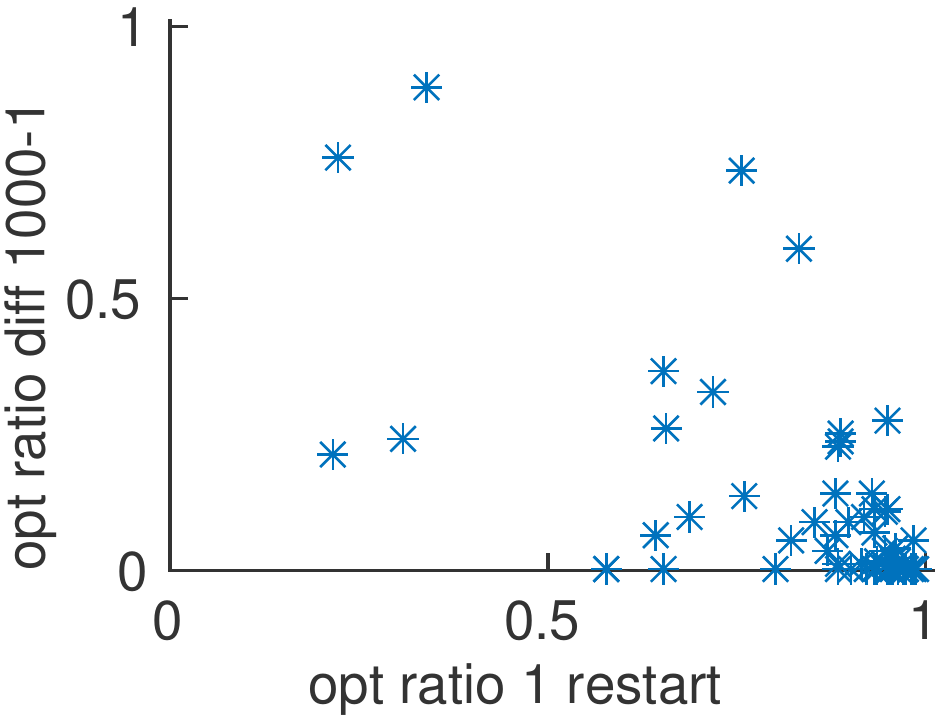} 
\\ (d) & (e) & (f) \\
\end{tabular}}
\caption{ Results on synthetic two dimensional data. {\bf a.} An example with its generating Gaussians.  {\bf b. and c.} Mixtures estimated by projected EM. Both soulitions give high likelihood to the data but thanks to our convex upper bound we know that the one in {\bf b} is suboptimal while the one in {\bf c.} closer to the optimal.  {\bf d}  A subset from the discrete set of models (containing over 1 million models) constructed by exhaustive search. {\bf e.} A summary of the optimality ratio computed using our algorithm. Experiments were done over models with diferent number of mixture components (K). The ratio is plotted as a function of the c-separation of the mixture. {\bf f.} The correlation between the optimality ratio after 1 restart of projected EM and the improvement in the optimality ratio after 1000 restarts. Our convex upper bound allows us to assess whether additional restarts are needed.} 
\label{results-fig1}
\end{figure*}

We first performed experiments with synthetic two dimensional data. 
In low dimensions, one
can perform {\em exhaustive enumeration} over mixture parameters. A
two dimensional Gaussian is defined by 5 numbers (two coordinates of
the mean, two eigenvalues of the covariance, and one rotation for the
covariance). We used 3,000 possible means, $12$ possible eigenvalues
and $8$ possible angles to create a set of one million possible
Gaussians. We generated data from finite mixtures whose parameters where chosen randomly (and hence not part of the discrete set) and ran convex EM to obtain an upper bound as well as projected EM to obtain lower bounds. Figure~\ref{results-fig1}a shows an example of a generated data and figure~\ref{results-fig1}d shows a subset of the discrete set of models.

 Figure~\ref{results-fig1}b-c show two solutions found by EM for this dataset. Both solutions give high likelihood to the data but our convex upper bound allows us to say that one is subpotimal (achieves only 90\% optimality ratio) while the other is closer to optimal (96\% optimality ratio). Without our bound, there would be no easy way to tell how suboptimal the solution in figure~\ref{results-fig1}b is. 
Figure~\ref{results-fig1}e  shows 
 the optimality ratio between the log
likelihood found by EM and our computed upper bound on the
optimal solution. We plot these ratios as a function of the separation
of the mixture Gaussians ~\cite{Dasgupta}.  A 2-separated mixture corresponds
 roughly to almost completely separated Gaussians, whereas a mixture
 that is $1$ or $\frac{1}{2}$-separated contains Gaussians which overlap
 significantly \cite{Dasgupta}. As can be seen, in many cases the optimality ratio is close to 100\%. Thus our convex upper bound calculation allows us to prove that we have found the global optimum of the log likelihood for these problems.  Figure~\ref{results-fig1}f shows the improvement in optimality ratio obtained when we run EM with 1000 random restarts as compared to a single run. This improvement is plotted as a function of the optimality ratio calculated after a single run of EM. As can be expected, when the optimality ratio is low, there is room for significant improvement by running EM with more restarts. Again, in the absence of a bound on the log likelihood it would be impossible to tell whether additional restarts will improve the likelohood or not.
 



\begin{figure*}
\centerline{
\begin{tabular}{cccc}
\includegraphics[width=0.23\textwidth]{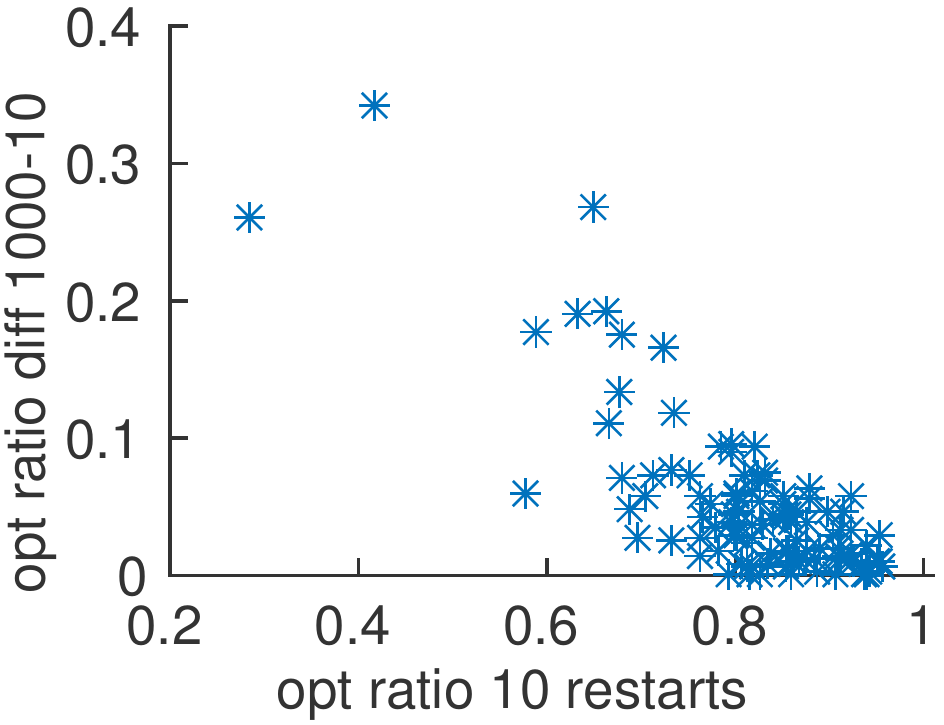} &
\includegraphics[width=0.23\textwidth]{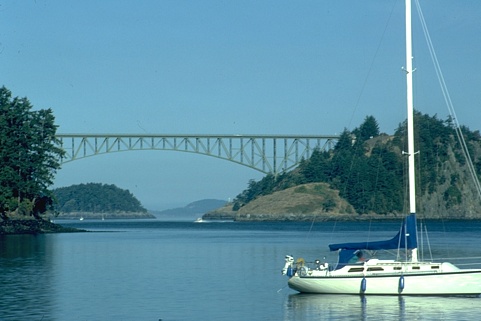} &
\includegraphics[width=0.23\textwidth]{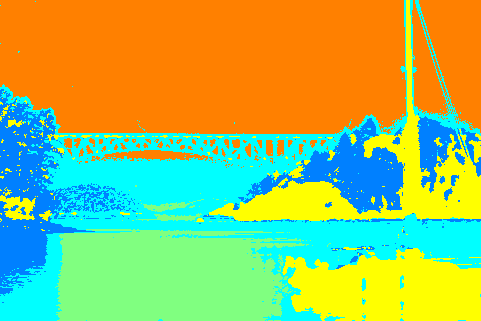} &
\includegraphics[width=0.23\textwidth]{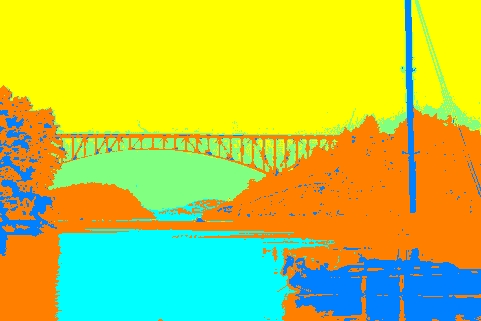}   
\\ (a) & (b) & (c) & (d) \\
\includegraphics[width=0.23\textwidth]{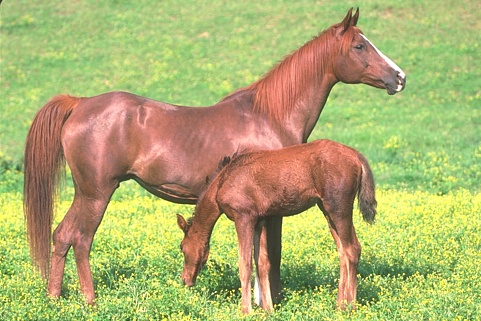} &
\includegraphics[width=0.23\textwidth]{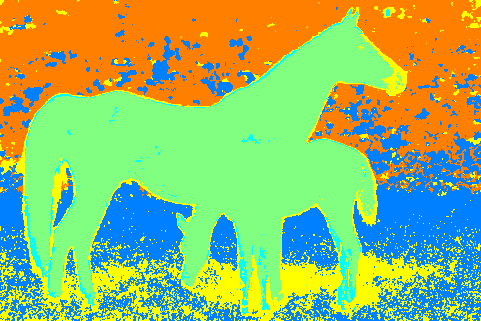} &
\includegraphics[width=0.23\textwidth]{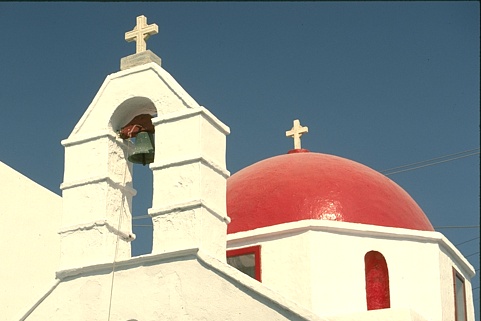} &
\includegraphics[width=0.23\textwidth]{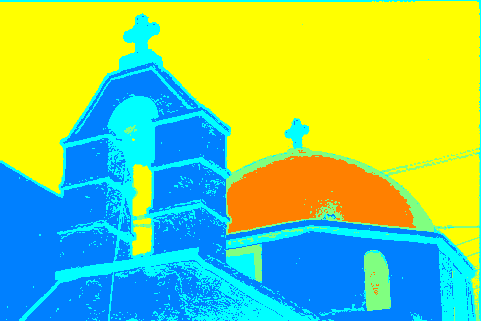} 
\\ (e) & (f) & (g) & (h) 
\end{tabular}}
\caption[]{{\bf a.} The correlation between the optimality ratio after 10 restarts of projected EM and the improvement in the optimality ratio after 1000 restarts over 100 images from BSD ~\cite{BSD} - when the optimality ratio is relatively low our benefit from more restarts would be greater. {\bf b,c,d.} An example where more restarts improved the optimality ratio (as well as the quality of the segmentation) from 85\% after 10 restarts ({\bf c}), to 90\% after 1000 restarts ({\bf d}) next to  the original image ({\bf b}). {\bf e,f,g,h.} More segmentatin results. Optimality ratio for image {\bf f} is 93\% and for image {\bf h} is 94\%.}
\label{seg-fig}
\end{figure*}

A popular use of mixture models in computer vision is in {\em image segmentation} where each segment is assumed to be modeled by a five dimensional Gaussian --- x,y and RGB (e.g.~\cite{goldberger}). Since this is a five dimensional problem exhaustive search for creating the set of models is infeasible. Instead 
we generate a discrete set of a  quarter of a million possible models by taking rectangular image patches of different sizes and robustly fitting a Gaussian to the pixels within that patch. Image segmentation is then performed by fitting a Gaussian mixture model with 5 components to the pixel data. 

We ran projected EM on the 100 images in the Berkeley Segmentaton Dataset~\cite{BSD}.  Figure~\ref{seg-fig}a shows a similar plot to what we saw in the synthetic data. When the optimality ratio is far from one, there is potential for significant improvement in the log likelihood values by rerunning EM. Figures~\ref{seg-fig}b-d show a particular example. After 10 restarts the optimality ratio is 85\% and so rerunning EM with multiple restarts improves the log likelihood significantly (and also leads to more natural looking segmentation). Figures~\ref{seg-fig}c-h show two examples where the bound allows us to assess how far we are from optimality.   Even though this data was not generated by a mixture of Gaussians, our convex upper bound allows us to give an optimality certificate for some of the images and also gives an indication regarding whether it is worthwhile to run EM with more restarts.

\section{Discussion}

Finite mixture models are widely used in machine learning and in
additional application areas. The standard method for estimating the
parameters of mixture models from data is the EM algorithm which is
known to sometimes find local optima. In this paper we have discussed
a method that can globally optimize the constrained maximum likelihood
problem: when each model's parameters come from a discrete set of
possible parameters 
(the models can have different forms, e.g. some could be Gaussians and others of uniform distribution). 
We have shown that an upper bound can be found
using a simple convex variant of EM and gave conditions under which
this bound is guaranteed to be tight. This bound can be used to assess
the performance of any estimation algorithm including projected EM.  

The strongest assumption made by our algorithm is that we can generate
a finite discrete set of models so that a mixture that only used
models inside the discrete set can approximate the true
distribution. In low dimensions this discrete set can be generated by
exhaustively covering a range of possible parameter values but in high
dimensions this is obviously intractable. Nevertheless our experiments
in image segmentation have shown that it
is often possible to generate this class of models by estimating
models based on subsets of the data.

Whenever optimization methods are applied to real-world problems and
the methods are not perfect, the question arises: are the failures due
to the optimization method used or to the cost function being
optimized? For discrete optimization methods (e.g. MAP in graphical
models), the answer to this question can often be obtained using the
linear programming relaxation (e.g.~\cite{RushSCJ10,WainwrightJ08}) which
gives a problem-dependent bound on the optimal value of the
optimization problem. To the best of our knowledge, our paper is the
first to give an analogous bound for the problem of maximizing the
likelihood of a finite mixture model.

\section*{Acknowledgment}
This work has been supported by the the ISF. The authors wish to thank the anonymous reviewers for their helpful comments.

\bibliography{rsemrefs}
\bibliographystyle{plain}



\end{document}